%% file: main.tex
\documentclass[letterpaper, 10 pt, conference]{ieeeconf}  % Comment this line out
\IEEEoverridecommandlockouts                              % This command is only
\usepackage[utf8]{inputenc}
\usepackage[T1]{fontenc}
\usepackage{mathptmx} % assumes new font selection scheme installed
\usepackage{mathptmx} % assumes new font selection scheme installed
\usepackage{amsmath} % assumes amsmath package installed
\usepackage{amssymb}  % assumes amsmath package installed
\usepackage{graphicx} 
\usepackage{grffile}
\usepackage{silence}  % silences warnings (specifically because caption does not recognize the ieeeconf document class
\usepackage[hypcap=true]{caption}
\usepackage{array}
\usepackage{xcolor}
\usepackage{mathtools}
\usepackage{commath}
\usepackage{colortbl}
\usepackage{enumerate}
\usepackage{hyperref}
\usepackage{pifont}

\usepackage{float}
\usepackage{tabularx}
\usepackage{makecell}
\usepackage{multirow}
\usepackage{placeins}
\usepackage{hyperref}
\usepackage{gensymb}
\usepackage{tabu}
\usepackage{cite} % gg added
\usepackage{balance}
\usepackage{tikz}
\usepackage{pgfplots}
\usepackage{subfig}

\newcommand{\jwg}[1]{}
\newcommand{\eva}[1]{}
\newcommand{\gok}[1]{}
\newcommand{\change}[1]{}

\title{\LARGE \bf
Standing Tall: Sim to Real Fall Classification and Lead Time Prediction for Bipedal Robots 
}
\author{ Gokul Prabhakaran, Jessy W. Grizzle, and M. Eva Mungai %
\thanks{\{Robotics, Robotics, Robotics\} Department, University of Michigan, Ann Arbor, MI, USA.{\text{\{gok,grizzle, mungam\}@umich.edu}}}
}

\begin{document}
\maketitle
\thispagestyle{empty}

\begin{abstract}
This paper extends a previously proposed fall prediction algorithm to a real-time (online) setting, with implementations in both hardware and simulation. The system is validated on the full-sized bipedal robot Digit, where the real-time version achieves performance comparable to the offline implementation while maintaining a zero false positive rate, an average lead time (defined as the difference between the true and predicted fall time) of 1.1s (well above the required minimum of 0.2s), and a maximum lead time error of just 0.03s. It also achieves a high recovery rate of 0.97, demonstrating its effectiveness in real-world deployment. In addition to the real-time implementation, this work identifies key limitations of the original algorithm, particularly under omnidirectional faults, and introduces a fine-tuned strategy to improve robustness. The enhanced algorithm shows measurable improvements across all evaluated metrics, including a 0.05 reduction in average false positive rate and a 1.19s decrease in the maximum error of the average predicted lead time.
\end{abstract}

% *********************  BEGIN PAPER  ***********************

% Introduction/Motivation
\input{sections/introduction}
\input{sections/hardware_description}
\input{sections/online_results_nominal}
\input{sections/analysis_nominal}
\input{sections/fine-tuned}
\input{sections/conclusion}
\balance
\bibliographystyle{template/IEEEtran}
\bibliography{references.bib}
\end{document}

%% file: sections/introduction.tex
\section{Introduction}

Humanoid robots have attracted significant attention over the past year, with a surge in companies developing general-purpose humanoids for home and factory applications. Although some humanoids have already been deployed in home and factory environments \cite{ref:morgan_stanley_report} and their performance in the recent World Humanoid Robot Games in Beijing showcased their increased ability to handle complex situations\cite{ref:wall_street_journal}, their widespread adaptation, especially in unstructured or unknown environments, remains limited. A key challenge in broadly deploying humanoid robots in these environments is their high-dimensional, hybrid nature, which, paired with their smaller support polygons, makes it particularly difficult to maintain stable motion, especially in unstructured terrains in the presence of disturbances and uncertainties. Although a well-designed controller can mitigate the effects of disturbances and uncertainties, the unpredictable nature of real-world operations makes falls inevitable. This is especially critical in applications involving human-robot interaction where safety is paramount.  As a result, it is imperative to implement fall prediction algorithms alongside robust controllers. Therefore, this paper extends the previously proposed fall prediction framework \cite{ref:mungai2024fall} to real-time operations, analyzes the algorithm's robustness to omnidirectional disturbances during the task of standing, and offers pathways to expand its capabilities. Standing was chosen in \cite{ref:mungai2024fall} as a precursor to more dynamic behaviors.   

\subsection{Background}
The overarching objective of a fall prediction algorithm is to accurately predict falls with sufficient time to implement a recovery strategy. As discussed in \cite{ref:mungai2024fall} and \cite{ref:mungai2023optimizing}, faults can be a precursor to a bipedal robot falling. Faults are defined as unforeseen deviations in one or more operational variables. Based on their temporal behavior, they can be categorized as either \textbf{abrupt}, \textbf{incipient}, or \textbf{intermittent}. Abrupt faults are sudden, incipient faults develop gradually, and intermittent faults occur sporadically. Intermittent faults are comprised of abrupt and incipient faults \cite{ref:ISERMANN19827}. In this paper, we refer to faults that lead to a fall as \textbf{critical faults}. The term \textbf{lead time} is used to describe the difference in time between the prediction of a fall and its actual occurrence, while \textbf{response time} is the difference in time between when a critical fault is introduced and when a fall is predicted. A trajectory or state is considered \textbf{unsafe} if it leads to a fall, and \textbf{safe} if it does not. 

\begin{figure}[t]
    \centering
    \begin{minipage}{0.4\textwidth}
        \centering
        \includegraphics[width=\textwidth]{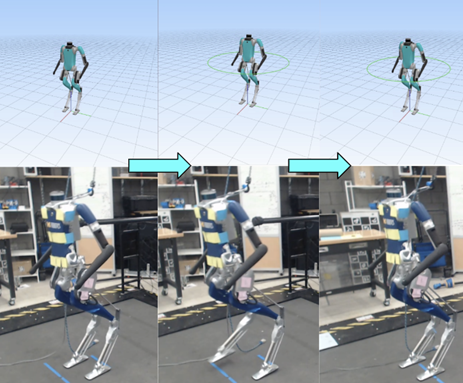}
        \caption{Real-time detection of a critical fault introduced between the first and second panels.}
        \label{fig:sim_hdw_switch_recovery}
    \end{minipage}
\end{figure}

\subsection{Literature Review}

While numerous fall prediction algorithms have been proposed for bipedal robots—ranging from static threshold-based methods to machine learning approaches \cite{ref:subburaman2023survey, ref:new_zhang2024fall}—the implementation of algorithms capable of real-time prediction of falls caused by abrupt, incipient, or intermittent faults in full-sized humanoid robots remains limited. Moreover, the prediction of lead time is also limited. Given the trade-off between false positive rate and lead time, and the dependence of the minimum required lead time on both the recovery algorithm and hardware platform, lead time prediction can significantly inform the selection and timing of an appropriate recovery strategy.

Fall prediction algorithms in the literature can be categorized based on several factors: the type of faults they address (abrupt, incipient, or intermittent), the size of the robot used, the experimental setting (offline with simulation data, offline with hardware data, online in simulation, or online in hardware), and whether they estimate lead time. In this section, we briefly go over the fall prediction algorithms in the literature based on these categorizations.

The fall prediction algorithm proposed in \cite{ref:subburaman} uses sensor fusion to predict falls in real-time, across various terrains and during dynamic motions such as squatting, in simulation and hardware for the 1.92m WALK-MAN humanoid \cite{ref:walkman}. However, the analysis is limited to falls caused by abrupt faults. Similarly, the SVM-based algorithm from \cite{ref:wu2021falling} is tested online in hardware and simulation for the 1.6m BHR-6 robot during walking and standing tasks, and only considers abrupt faults. 

In contrast, while \cite{ref:new_igual2024time}, \cite{ref:tran}, \cite{ref:khaled_2025}, and \cite{ref:ZHANG2025104995} do not explicitly address all three faults, their analyses are based on scenarios that may encompass all temporal fault categories—for instance, varying the robot's initial state \cite{ref:new_igual2024time}, robot-to-robot interactions \cite{ref:tran}, walking on uneven terrain \cite{ref:khaled_2025}, and motor failures \cite{ref:ZHANG2025104995}. Despite the potential inclusion of all three faults, the analyses in \cite{ref:new_igual2024time} and \cite{ref:khaled_2025} are limited to offline simulations of the 1.43m planar bipedal robot Rabbit \cite{ref:chevallereau2003rabbit} and the 90cm NUgus humanoid robot, respectively. While \cite{ref:tran} conduct their analysis online in both hardware and simulation, it is limited to the 0.57m tall Nao H25 robot \cite{ref:nao, ref:naoh25}. On the other hand, \cite{ref:ZHANG2025104995} uses both hardware and simulation data to evaluate their algorithm on the 1.82m Q1 humanoid robot and a Cassie bipedal robot—originally 1m tall\cite{ref:gong2019feedback}—fitted with a torso, however, their analysis does not include online hardware experiments. 

\cite{ref:mungai2024fall} proposes a 1D CNN model capable of predicting falls caused by abrupt, incipient, and intermittent faults, both in simulation and hardware, using the 1.6m tall bipedal robot, Digit \cite{ref:ar_digit}. While comparable to \cite{ref:ZHANG2025104995}, the algorithm in \cite{ref:mungai2024fall} is, to the best of our knowledge, the only one that also estimates lead time on a full-sized humanoid. Additionally, the authors provide an open-source dataset for reproducibility \cite{ref:digit_dataset}. However, the study is limited to offline prediction in a planar standing task, where faults are simulated by applying external forces to the robot's back.

\subsection{Objective}
Given the high offline accuracy in critical fault detection and lead time prediction of the fall prediction algorithm proposed by \cite{ref:mungai2024fall}, along with the availability of its dataset, the objective of this paper is to extend the algorithm to a real-time (online) setting and validate its performance in both hardware and simulation on the full-sized bipedal robot Digit. Furthermore, as the algorithm is only trained and evaluated on faults introduced by horizontal forces applied to the center back of the robot's torso, this work seeks to identify its limitations under omnidirectional faults during a standing task and propose a fine-tuned strategy to improve robustness with minimal additional data. Note that the task of standing is chosen as a precursor to more dynamic motions. The algorithm proposed in \cite{ref:mungai2024fall} is referred to as the \textbf{nominal algorithm}.

Specifically, this paper:
\begin{itemize}
    \item Maintains the nominal algorithm’s goal of reliably detecting imminent falls caused by abrupt, incipient, and intermittent faults, with sufficient lead time for a full-sized bipedal robot, while also accurately predicting lead time.
    \item Adopts the nominal fall prediction algorithm to an online setting, both in hardware and in simulation.
    \item Evaluates the performance of the nominal algorithm on critical faults applied in the sagittal and frontal plane at various angles and locations on the robot, using the same metrics: false positive rates, lead time, and response time.
    \item Proposes a strategy for robustifying the nominal algorithm against a broader set of faults.
\end{itemize}

As in \cite{ref:mungai2024fall}, the robot is considered to have fallen if its center of mass is less than 10 percent of its initial height, and the minimum desired lead time is set to 0.2s. 

Achieving the objective outlined above is challenging due to several factors: the masking effects of the controller, the sporadic nature of intermittent faults, the trade-off between lead time and false positive rates, and the crowding effect of incipient faults \cite{ref:safaeipour2021survey}. Furthermore, the decreasing number of available data points as the lead time increases \cite{ref:mungai2024fall}, along with the diverse range of faults, complicates the prediction of falls. Implementing the algorithm online introduces additional challenges, including the need for fast inference times and the management of communication delays between different components of the fall prediction algorithm and other processes, such as the controller.

\subsection{Contributions}
The main contributions of this paper are as follows:
\begin{itemize}
    \item Implementation of the nominal fall prediction algorithm online, both in hardware and simulation
    \item Extension of the open-source dataset \cite{ref:digit_dataset} to include critical and non-critical faults, introduced by applying angled and non-angled forces at various locations in the sagittal and frontal planes of the robot
    \item Analysis and robustification of the nominal fall prediction algorithm against a broader range of faults.
\end{itemize}
While our study focuses on the nominal fall prediction algorithm for the bipedal robot Digit during standing, the techniques used are applicable to other algorithms, platforms, and tasks.

%% file: sections/hardware_description.tex
\section{Method: Nominal Fall Prediction Algorithm, Hardware Description}
This section provides a brief overview of the nominal fall prediction algorithm, along with a description of the robot hardware used.

\subsection{Overview of Nominal Fall Prediction Algorithm Framework}
\label{subsec:prev_framework}
\begin{figure}
\vspace{2mm}
    \centering
    \includegraphics[width=0.85\linewidth]{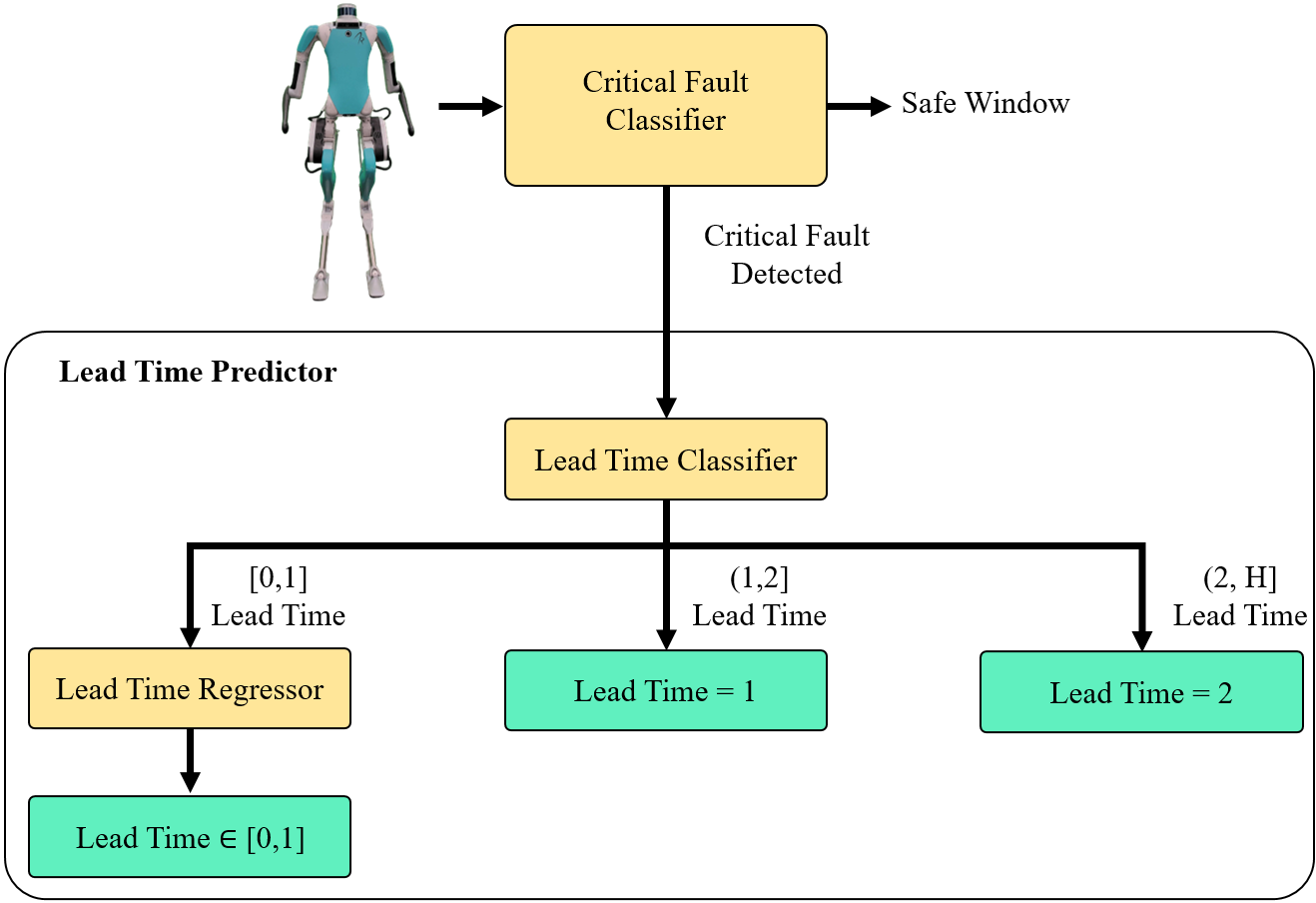} % GG edit
    \caption{The nominal fall prediction algorithm is comprised of three components: a critical fault classifier, a lead time classifier, and a lead time regressor. The green boxes depict the algorithm's predicted lead time.}
    \label{fig:fall_prediction_algorithm}
    \vspace{-2mm}
\end{figure}

The 1D CNN-based nominal fall prediction algorithm consists of three parts: a critical fault classifier, a lead time classifier, and a lead time regressor. The critical fault classifier identifies critical faults, while the lead time classifier and the regressor work together to predict the lead time. The lead time classifier divides the lead time into three intervals: $[0,1]$, $(1,2]$, and $(2, H]$. If the lead time is classified into the $[0,1]$ interval, the lead time regressor calculates the lead time. Otherwise, the lead time is considered as the infimum of the interval. To prioritize the most recent data points, sliding windows with 10 data points are used. The algorithm is trained with trajectories sampled at approximately 40 Hz. Fig. \ref{fig:fall_prediction_algorithm} depicts the algorithm's framework.

\subsection{Hardware Description}
\begin{figure}
\vspace{2mm}
    \centering
    \includegraphics[scale=0.45]{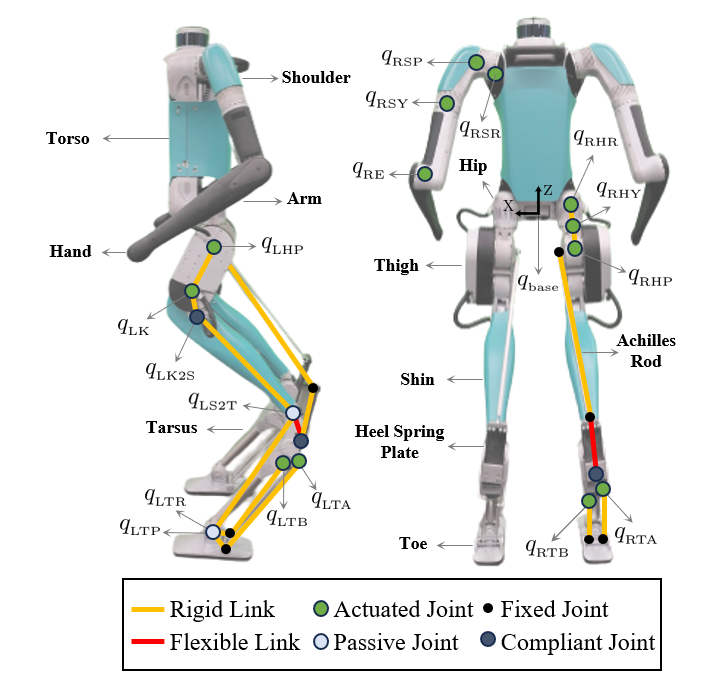}
    \caption{Kinematics architecture of the Digit robot by Agility Robotics \cite{ref:ar_digit, ref:mungai2024fall}.}
    \label{fig:digitHardware}
    \vspace{-2mm}
\end{figure}

We utilize the bipedal robot Digit \cite{ref:ar_digit} developed by Agility Robotics. Digit is approximately 1.6m, weighs 48 kg and has 30 degrees of freedom, 20 actuated joints, and an integrated perception system. Fig. \ref{fig:digitHardware} illustrates Digit's kinematic architecture. 

%% file: sections/online_results_nominal.tex
\section{Online Implementation and Evaluation of Nominal Fall Prediction Algorithm}
\label{sec:online_fall_pred}
This section presents the implementation of the nominal fall prediction algorithm online in both simulation and hardware and assesses it's performance.

\subsection{Online Implementation}
As in the offline implementation, we employ the use of Agility's MuJoCo-based simulator with a standing controller designed to maintain the center of mass and zero moment point inside the support polygon \cite{ref:grant}. This standing controller will be referred to as the \textbf{nominal controller}. To run the fall prediction algorithm online, the robot is first initialized in a standing position with the nominal controller. The robot's state and the torques commanded by the nominal controller are then queried via ROS \cite{ref:ros} at a rate of 400Hz. The nominal fall prediction algorithm, however, runs at 40 Hz. Therefore, features for the algorithm are computed using newly queried data every 40 Hz. Once enough features are available to form a window, model inference is performed continuously until a fall is predicted. Once a fall is predicted, the recovery controller, if one is available, is activated. The average runtime of the nominal fall prediction algorithm is approximately 0.0045s, with the total inference time of the 1D CNN models being 0.004s and the computation of the features taking 0.0005s. 

\subsection{Evaluation Metrics}
The algorithm's performance online, following the offline analysis in \cite{ref:mungai2024fall}, is evaluated using lead time, false positive rate, and response time. In addition to these metrics, we introduce a new metric to assess the algorithm's effectiveness: the \textbf{recovery rate}. The recovery rate measures the number of unsafe trajectories that are stabilized by switching to a recovery controller when a critical fault is detected. We report the recovery rate only for simulation tests due to the difficulty of replicating the same fault multiple times on the hardware. Since implementing a recovery controller falls outside the scope of this work, we utilize Agility's base controller, which already incorporates a fall prevention strategy. We note that only abrupt and incipient faults are considered in our evaluation, as the nominal fall prediction algorithm's effectiveness in detecting intermittent faults, as demonstrated in \cite{ref:mungai2024fall}, depends on accurately identifying critical abrupt and incipient faults.

\subsection{Experimental Setup}
Abrupt and incipient faults are emulated online in simulation by applying impulsive and trapezoidal pushing forces, respectively, to regions F in Fig. \ref{fig:digit_push_locations}. To assess the fall prediction algorithm's performance online in simulation and compare it to the offline results presented in \cite{ref:mungai2024fall}, we replicate the 200 abrupt and incipient tests used in the offline evaluation and apply the algorithm to these tests. The simulation is run on a machine with an Intel i9-7900x processor.

In the hardware setup, we do not have direct access to the main computer, so we interact with the robot by sending commands and receiving state information over UDP at approximately 1 kHz. This information is processed by the nominal controller, which publishes topics via ROS also at roughly 1 kHz. The robot is connected to an Alienware m15 computer with an Intel i9-12900H processor via Ethernet. This computer runs both the fall prediction algorithm and the nominal controller.

We execute the algorithm online in hardware 13 times, emulating 8 abrupt faults and 6 incipient faults. In contrast to the simulation setup, faults are introduced by applying pushing forces using a pole to region C in Fig. \ref{fig:digit_push_locations}, as region F houses the on-board computer. To capture the gradual nature of incipient faults, the pole is first positioned on the robot before applying force. To prevent potential damage during experiments, the robot is secured to a gantry using a slack cable. Note that the nominal fall prediction algorithm is transferred from hardware to simulation in a zero-shot manner - no additional data is used for training, and no modifications are made to the algorithm.

\subsection{Performance Analysis}
\subsubsection{Simulation — Offline vs Online}
\begin{figure}[htbp]
\vspace{2mm}
    \centering
    \begin{minipage}{0.48\textwidth}
        \centering
        \includegraphics[width=\textwidth]{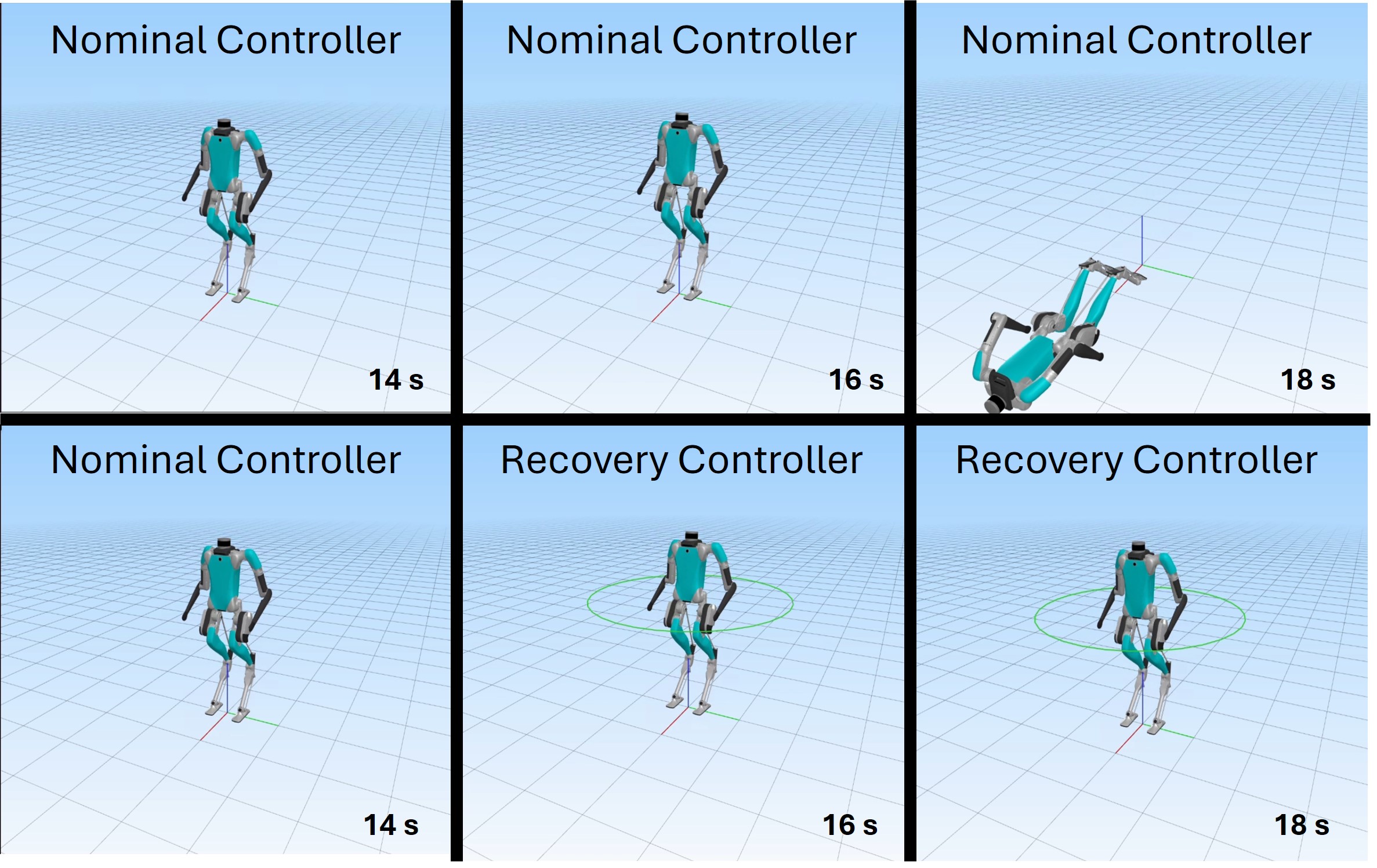}
        \caption{Showcases the results of switching vs not switching to a recovery controller when the nominal fall prediction algorithm correctly detects a critical abrupt fault. The fault is introduced at 15s.}
        \label{fig:sim_switch_recovery}
    \end{minipage}
    \vspace{-4mm}
\end{figure}

As shown in Table \ref{tab:on_vs_off_sim_results}, the true average lead time for the online execution in simulation is 0.01s higher than the offline execution. This minor discrepancy in fault detection is primarily due to communication delays, which cause the online algorithm to occasionally run slower, averaging 40 Hz but sometimes dipping to lower rates. As a result, windows with features are sampled at intervals slightly greater or less than 40 Hz, leading to differences in the timing of critical fault detection. These timing variations also contribute to small changes in the average response time and predicted lead time, which decrease by 0.02s and increase by 0.04s, respectively, compared to the offline results. The difference between the average true lead time and predicted lead time is 0.03s for the online execution and 0.02s for the offline execution. Both online and offline evaluations report an average false positive rate of 0.  Given the small differences between the online and offline results, we can conclude that the algorithm's online performance in simulation is closely aligned with its offline results. Furthermore, the 0.97 recovery rate for the unsafe trajectories demonstrates the effectiveness of running the fall prediction algorithm online. Fig. \ref{fig:sim_switch_recovery} showcases the results of switching vs not switching to a recovery controller when the nominal fall prediction algorithm detects a critical fault.

\begin{table}[htbp]
\centering
\caption{Online vs Offline Average Simulation Results of the Nominal Fall Prediction Algorithm Evaluated on Abrupt and Incipient Faults }
\begin{tabular}{|c|c|c|c|c|c|}
\hline
{\makecell{Operation}} & \makecell{Predicted\\Lead\\Time }  & \makecell{True\\Lead\\Time}  & {\makecell{False\\Positive\\Rate}} & {\makecell{Response\\Time\\ (s)}} & {\makecell{Recovery\\Rate}} \\ \hline
Offline\cite{ref:mungai2024fall} & 1.18 & 1.16 & 0.0 & 0.44 & N/A \\ \hline
Online  & 1.14 & 1.17 &  0.0 & 0.42 & 0.97 \\ \hline 
\end{tabular}
\label{tab:on_vs_off_sim_results}
\end{table}

\subsubsection{Hardware — Offline vs Online}
In the hardware tests online, 4 of the 8 abrupt faults are detected as critical faults, and 3 of the 6 incipient faults are identified as critical faults. The average predicted lead time is 1.04s which is similar to the offline hardware results reported in \cite{ref:mungai2024fall}. Note that we only report the average predicted lead time, as measuring the force application time and hence the response time online without force sensors proved challenging. Additionally, false-positive rates are unknown since the recovery controller is triggered whenever falls are detected. However, the similarity in average predicted lead times between the online and offline implementations indicates that the nominal fall prediction algorithm’s performance on hardware remains consistent with its offline results and the simulation results. Fig. \ref{fig:hdw_switch_recovery} demonstrates the fall prediction algorithm detecting a critical fault online in hardware.

\subsubsection{Simulation vs Hardware — Online}
\begin{figure}
\vspace{2mm}
    \centering
    \begin{minipage}{0.48\textwidth}
        \centering
        \includegraphics[width=\textwidth]{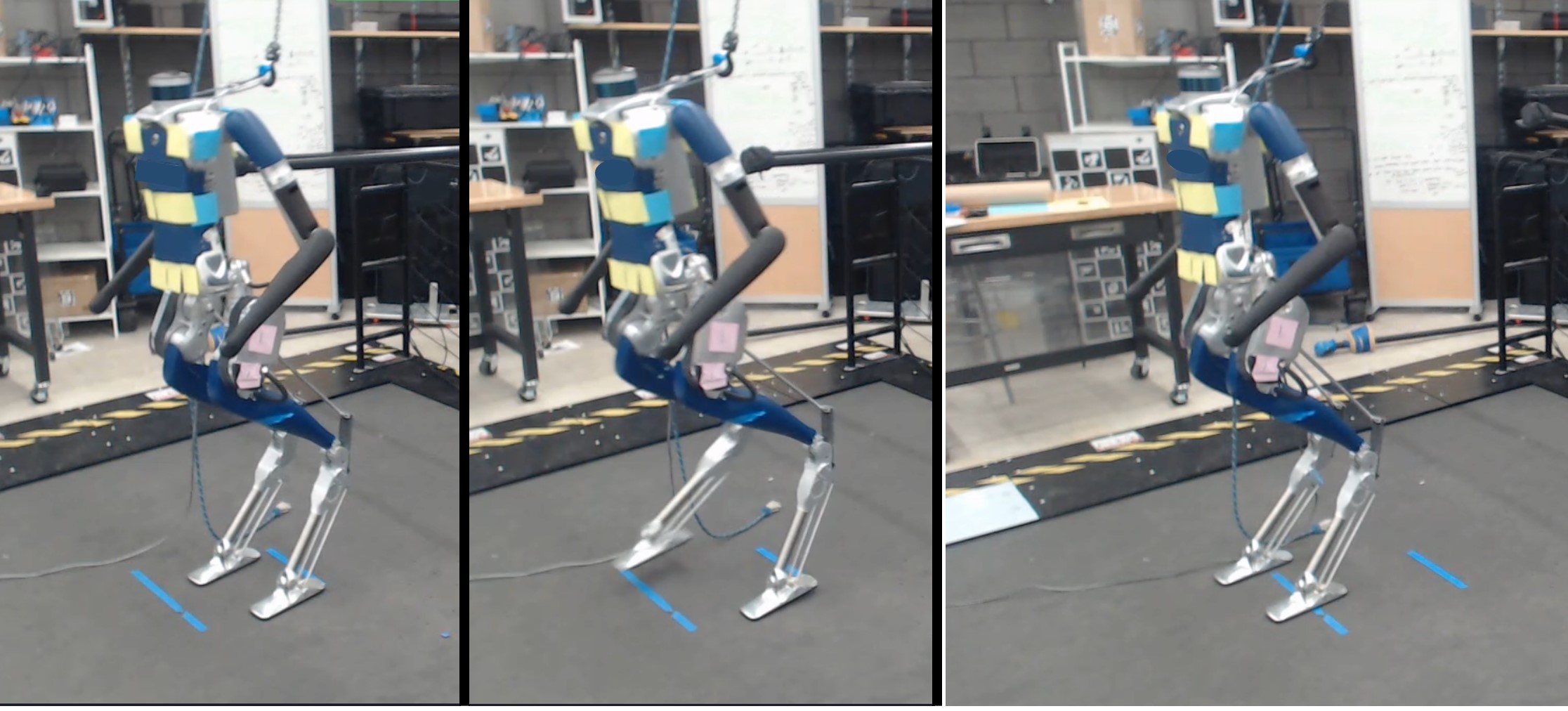}
        \caption{The nominal fall prediction algorithm detects a critical incipient fault and switches to the recovery algorithm}
        \label{fig:hdw_switch_recovery}
    \end{minipage}
    \vspace{-2mm}
\end{figure}

The average predicted lead time achieved online in hardware is 0.1s lower than the results obtained in the online simulation. This deviation can be attributed to the smaller number of hardware online tests, which do not encompass as broad a range of critical faults. In contrast, the average runtime of the fall prediction algorithm for the hardware setup is 0.0061s, which is comparable to the algorithm's runtime in simulation. The 0.0016s difference in runtime is negligible, given the algorithm's sampling rate of 0.025s (40Hz). Furthermore, the nominal controller publishes at about 1kHz in both hardware and simulation. The negligible difference in runtime for the fall prediction algorithm between simulation and hardware, the small sim-to-real gap demonstrated offline in \cite{ref:mungai2024fall}, and the similarity in performance when executed offline and online in simulation, as shown in Table \ref{tab:on_vs_off_sim_results}, suggest that the algorithm's performance offline in simulation reliably translates online to hardware, further validating its real-world applicability. Fig. \ref{fig:sim_hdw_switch_recovery} offers a consolidated view of critical fault detection in both hardware and simulation, with the recovery algorithm in the loop, as illustrated in Figs. \ref{fig:sim_switch_recovery} and \ref{fig:hdw_switch_recovery}.

%% file: sections/analysis_nominal.tex
\section{Analysis of Nominal Fall Prediction Algorithm on Omnidirectional Faults}
%%this section should describe how the nominal classifier, regressor, and bin classifier perform offline under various faults (e.g. angled and applied at different parts of the robot)

Following the impressive online performance of the nominal fall prediction algorithm, we now evaluate its robustness to omnidirectional faults. The nominal fall prediction algorithm was trained on both abrupt and incipient faults introduced by applying pushing forces to region F at the back of the torso, as shown in Fig. \ref{fig:digit_push_locations}, using features defined exclusively within the sagittal plane. As such, it is not expected to generalize well to faults outside of its training distribution, such as those introduced by pushing the robot on its sides \cite{DBLP:journals/corr/HendrycksG16c}. However, since the training set included several trajectories in which the robot exhibited rocking motions before falling forward or backward within the sagittal plane, it remains an open question whether the algorithm can adapt to previously unseen fault conditions, such as those caused by forces applied at different locations and angles. To address this, we evaluate the performance of the three components of the fall prediction algorithm across a broad range of faults. We limit our analysis here to abrupt and incipient faults for reasons similar to those discussed in Section \ref{sec:online_fall_pred}. In addition, the evaluation is performed offline. However, given the negligible difference in performance between online and offline execution, as presented in Section \ref{sec:online_fall_pred}, we can reasonably conclude that the offline results presented here are indicative of the algorithm's performance in real-time, online scenarios. Furthermore, the results in \cite{ref:mungai2024fall} show that the performance of the individual components directly translates to the overall performance of the algorithm. Note that the Digit open-source dataset, \cite{ref:digit_dataset}, will be extended to include the trajectories used here.

\subsection{Data Generation of Omni-directional faults}
\label{sec:data_gen_angled}
\begin{figure}
\vspace{2mm}
    \centering
    \includegraphics[width=0.8\linewidth]{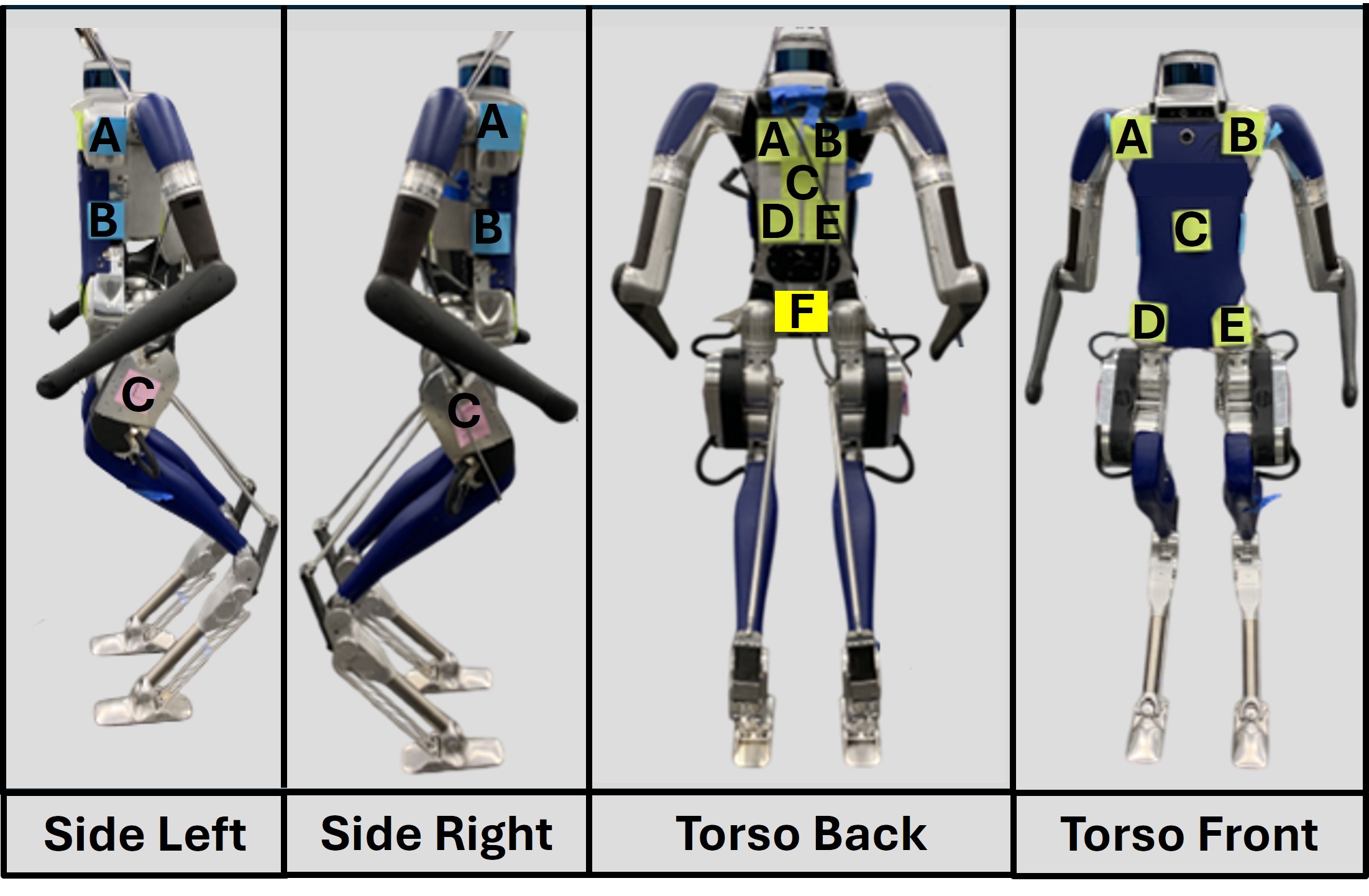} % GG edit
    \caption{Stickers in 16 positions indicating where faults were introduced. This includes (leftmost) 3 positions on the left side, (2nd from left) 3 positions on the right, (2nd from the right) 5 positions on the back of the torso, and (rightmost) 5 positions on the front of the torso. \eva{Gokul: Could you please update this figure.}}
    \label{fig:digit_push_locations}
    \vspace{-2mm}
\end{figure}

To assess the algorithm's robustness across a wider range of faults, abrupt and incipient faults are emulated using pushing forces as described in Section \ref{sec:online_fall_pred}. The test dataset is generated by applying both angled and non-angled forces at 16 distinct locations on the robot, distributed across the sagittal and frontal planes. These locations are grouped into four regions: torso back, torso front, right side, and left side, with five points on the back, five on the front, and three on each side. Fig. \ref{fig:digit_push_locations} illustrates the regions and points where the forces are applied.  

For each of the 16 locations, 100 trajectories are generated for both abrupt and incipient faults, resulting in 200 trajectories per location, using both angled and non-angled pushes. For convenience, we term critical faults that are introduced by angled and non-angled forces as critical angled and non-angled faults, respectively. Angled forces are applied by selecting the magnitudes of the x and y components of the force, as well as the angle of the force relative to the transverse plane of the robot. The x-component magnitude is chosen such that applying a non-angled x force causes a fall in approximately half of the trajectories (for both the torso back and front). Similarly, the y-component magnitude is set to produce a fall in about half of the trajectories when non-angled y forces are applied (for the torso sides). The angle of the force relative to the transverse plane is randomly selected from a uniform distribution within the range of [-45°, 45°]. These angles were determined through empirical testing, where the physical robot was pushed at various angles to identify the range within which it could be effectively disturbed. To ensure slight oscillations are present in the robot's standing posture, random impulsive forces are introduced, but only windows during and after the fault introduction are used for fall prediction.

\subsection{Nominal Critical Fault Classifier Results}
\label{sec: limitations_of_classifier}
% The objective of the critical fault classifier is to predict critical faults while maximizing lead time and minimizing false positive rates. 
In \cite{ref:mungai2024fall}, the critical fault classifier demonstrated impressive results by detecting abrupt and incipient faults, achieving an average lead time of 1.52 seconds, an average response time of 0.44 seconds, and an average false positive rate of 0 when evaluated on unsafe trajectories derived in a similar manner to its training set. We refer to these results as \textbf{in-distribution results}.

When assessed on the broader range of abrupt and incipient critical faults, the classifier achieves an average response time comparable to the in-distribution results, as shown in Table \ref{tab:classifier_on_normal_and_angled_pushes}. While its average lead times decrease by 0.17s and 0.09s for non-angled and angled critical faults, respectively, the corresponding average false positive rates also rise to 0.08 for both cases. This indicates a decline in the nominal critical fault classifier's performance. This decline in performance is particularly apparent for side pushes, as the false positive rates for critical faults applied to the sides of the robot are on average 0.13 for both angled and non-angled critical faults, whereas for critical faults introduced to the torso, the rates are 0.02 and 0.04 for non-angled and angled forces, respectively. These results align with expectations, as the model's training data primarily focuses on faults introduced along the sagittal plane and does not adequately capture the variability of side or angled faults.

\begin{table}[h]
\centering
\begin{tabular}{|c||c|c|c||c|c|c|}
\hline
 & \multicolumn{3}{c||}{\textbf{Nominal}} & \multicolumn{3}{c|}{\textbf{Fine-Tuned}} \\ \hline
\textbf{Metrics} & {\makecell{Non\\ Ang}} & {\makecell{Ang.}}  & {\makecell{In\\Dist}} & {\makecell{Non\\Ang}} & {\makecell{Ang}}  & {\makecell{In\\Dist}} \\ \hline
\makecell{FPR}              & 0.08 & 0.08 & 0.0  & 0.0  & 0.003 & 0.0 \\ \hline
\makecell{Lead\\Time}       & 1.35 & 1.43 & 1.52 & 1.18 & 1.26 & 1.18 \\ \hline
\makecell{Response\\Time}   & 0.42 & 0.41 & 0.44 & 0.59 & 0.59 & 0.62 \\ \hline
\end{tabular}
\caption{Nominal \cite{ref:mungai2024fall} and Fine-Tuned Critical Fault Classifier results evaluated on non-angled and angled abrupt and incipient faults introduced by forces applied at various locations on the robot. \textbf{Acronyms used: }
Non ang (non-angled), ang (angled), and in dist (in-distribution or tests from \cite{ref:mungai2024fall}). \label{tab:classifier_on_normal_and_angled_pushes}}
\end{table}

%%%%%%%%%%%%%%%%%%%%%%%%%%%%%%%%%%%%%%%%%%%%%%%%%%%%%%%%%%%%%%%%%%%%%%%%%%%%%%% 
\subsection{Nominal Lead time classifier}
\begin{figure}
\vspace{2mm}
    \centering
    \resizebox{0.865\linewidth}{!}{\input{figures/algorithm_analysis/old_lead_regressor_accuracy}}
    \caption{Lead time classifier accuracies when evaluated on non-angled and angled critical faults across various locations. The circles and squares indicate the non-angled and angled critical faults, respectively.}
    \label{fig:old_lead_time_classifier_accuracies}
    \vspace{-2mm}
\end{figure}
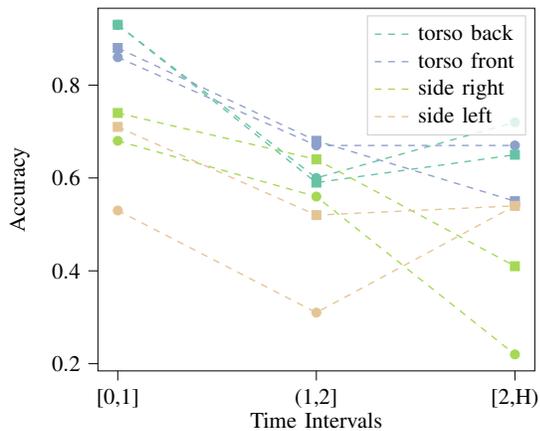

% The objective of the lead time classifier is to categorize lead times into three distinct intervals, $[0,1]$, $(1,2]$, and $(2,H]$. 
When evaluated on in-distribution critical faults, the lead time classifier achieved accuracies of 1.0, 0.95, and 0.74 for the $[0,1]$, $(1,2]$, and $(2,H]$ lead time intervals, respectively. However, similar to the critical fault classifier, the lead time classifier performs worse when evaluated on a broader range of critical faults. For non-angled critical faults, the classifier, as shown in Table \ref{tab:lead_time_classifier_results_on_normal_and_angled_pushes}, achieves accuracies of 0.82, 0.57, and 0.61 for the $[0,1]$, $(1,2]$, and $(2,H]$ intervals, respectively. For angled critical faults, the classifier achieves accuracies of 0.84, 0.63, and 0.55 for the same intervals. As shown in Fig. \ref{fig:old_lead_time_classifier_accuracies}, this decline in performance is particularly pronounced for side critical faults. These results align with those of the critical fault classifier and are consistent with expectations, especially considering that the lead time classifier achieves its highest accuracy within the $[0,1]$ interval. Recall that, as demonstrated in \cite{ref:mungai2024fall}, the number of available data points decreases exponentially as lead time increases. The asymmetry observed in Fig. \ref{fig:old_lead_time_classifier_accuracies} between the left and right sides can be attributed to randomness in data generation, leading to varying force magnitudes even within the same range. When critical faults from one side are emulated on the other, the lead time classifier produces similar results.

\begin{table}[h]
\centering
\begin{tabular}{|c||c|c|c||c|c|c|}
\hline
 & \multicolumn{3}{c||}{\textbf{Nominal}} & \multicolumn{3}{c|}{\textbf{Fine-Tuned}} \\ \hline
{\textbf{Intervals}} & {\makecell{Non\\Ang}} & {\makecell{Ang}}  & {\makecell{In\\Dist}} & {\makecell{Non\\Ang}} & {\makecell{Ang}}  & {\makecell{In\\Dist}} \\ \hline
\makecell{$[0,1]$} & 0.82 & 0.84 & 1.0 & 0.96 & 0.95 & 0.98 \\ \hline
\makecell{$(1,2]$} & 0.57 & 0.63 & 0.95 & 0.84 & 0.80 & 0.82 \\ \hline
\makecell{$[2,H)$} & 0.61 & 0.55 & 0.74 & 0.30 & 0.27 & 0.44 \\ \hline
\end{tabular}
\caption{Nominal \cite{ref:mungai2024fall} and Fine-Tuned Lead Time Classifier results evaluated on non-angled and angled abrupt and incipient faults introduced by forces applied at various locations on the robot.\textbf{ Acronyms used: }
Non ang (non-angled), ang (angled), and in dist (in-distribution or tests from \cite{ref:mungai2024fall}).}
\label{tab:lead_time_classifier_results_on_normal_and_angled_pushes}
\end{table}

%%%%%%%%%%%%%%%%%%%%%%%%%%%%%%%%%%%%%%%%%%%%%%%%%%%%%%%%%%%%%%%%%%%%%%%%%%%%%%%

\subsection{Nominal Lead time regressor}
\begin{figure}
    \centering
    \resizebox{0.865\linewidth}{!}{\input{figures/algorithm_analysis/old_regressor_results}}
    \caption{Lead time Regressor results when evaluated on non-angled and angled critical faults across various locations. The circles and squares indicate the non-angled and angled critical faults, respectively.}
    \label{fig:old_regressor_results}
\end{figure}
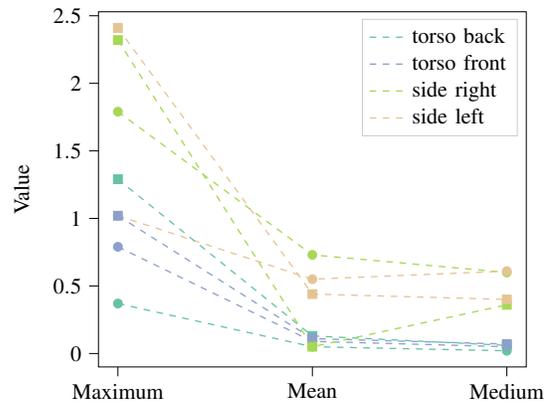

When evaluated on in-distribution critical faults within the $[0,1]$ lead time range, the lead time regressor shows minimal error, with a maximum difference of 0.09s between predicted and actual lead time, and a mean and median difference of just 0.01s \cite{ref:mungai2024fall}. However, similar to both the critical fault classifier and the lead time classifier, the regressor performs significantly worse on non-angled and angled critical faults. For non-angled critical faults, the maximum lead time prediction error increases to 1.79s, while the mean and median errors rise to 0.23s and 0.07s, respectively, as shown in Table \ref{tab:lead_time_regressor_results_on_normal_and_angled_pushes}. For critical angled faults, the maximum error increases to 2.41s, with mean and median errors of 0.25s and 0.11s, respectively. Given that the regressor's objective is to predict lead time within the $[0,1]$ interval, these error increases are unacceptable. Despite these significant errors, as shown in Fig. \ref{fig:old_regressor_results}, the regressor performs better for the torso back non-angled critical faults compared to the other regions and faults.

% This suggests that, like the lead time classifier and the critical fault classifier, torso front and back data align more closely with the regressor's training distribution.

\begin{table}[h]
\vspace{2mm}
\centering
\begin{tabular}{|c||c|c|c||c|c|c|}
\hline
 & \multicolumn{3}{c||}{\textbf{Nominal}} & \multicolumn{3}{c|}{\textbf{Fine-Tuned}} \\ \hline
{\textbf{Differences}} & {\makecell{Non\\Ang}} & {\makecell{Ang}}  & {\makecell{In\\Dist}} & {\makecell{Non\\Ang}} & {\makecell{Ang}}  & {\makecell{In\\Dist}} \\ \hline
\makecell{Max} & 1.79 & 2.41 & 0.09 & 0.23 & 0.32 & 0.18 \\ \hline
\makecell{Mean} & 0.23 & 0.25 & 0.01 & 0.03 & 0.03 & 0.04 \\ \hline
\makecell{Median} & 0.07 & 0.11 & 0.01 & 0.02 & 0.02 & 0.03 \\ \hline
\end{tabular}
\caption{Nominal and Fine-Tuned Lead Time Regressor results evaluated on non-angled and angled abrupt and incipient faults introduced by forces applied at various locations on the robot. \textbf{Acronyms used: }
Non ang (non-angled), ang (angled), and in dist (in-distribution or tests from \cite{ref:mungai2024fall}).}
\label{tab:lead_time_regressor_results_on_normal_and_angled_pushes}
\vspace{-2mm}
\end{table}

%% file: figures/algorithm_analysis/old_lead_regressor_accuracy.tex
% This file was created with tikzplotlib v0.10.1.
\begin{tikzpicture}

\definecolor{burlywood229196148}{RGB}{229,196,148}
\definecolor{darkgray141160203}{RGB}{141,160,203}
\definecolor{darkgray176}{RGB}{176,176,176}
\definecolor{lightgray204}{RGB}{204,204,204}
\definecolor{mediumaquamarine102194165}{RGB}{102,194,165}
\definecolor{yellowgreen16621684}{RGB}{166,216,84}

\begin{axis}[
legend cell align={left},
legend style={fill opacity=0.8, draw opacity=1, text opacity=1, draw=lightgray204},
tick align=outside,
tick pos=left,
x grid style={darkgray176},
xlabel={Time Intervals},
xmin=-0.1, xmax=2.1,
xtick style={color=black},
xtick={0,1,2},
xticklabels={{[0,1]},{(1,2]},{[2,H)}},
y grid style={darkgray176},
ylabel={Accuracy},
ymin=0.1845, ymax=0.9655,
ytick style={color=black}
]
\addplot [
  draw=mediumaquamarine102194165,
  fill=mediumaquamarine102194165,
  forget plot,
  mark=*,
  only marks
]
table{%
x  y
0 0.93
1 0.6
2 0.72
};
\addplot [draw=darkgray141160203, fill=darkgray141160203, forget plot, mark=*, only marks]
table{%
x  y
0 0.86
1 0.67
2 0.67
};
\addplot [draw=yellowgreen16621684, fill=yellowgreen16621684, forget plot, mark=*, only marks]
table{%
x  y
0 0.68
1 0.56
2 0.22
};
\addplot [draw=burlywood229196148, fill=burlywood229196148, forget plot, mark=*, only marks]
table{%
x  y
0 0.53
1 0.31
2 0.54
};
\addplot [
  draw=mediumaquamarine102194165,
  fill=mediumaquamarine102194165,
  forget plot,
  mark=square*,
  only marks
]
table{%
x  y
0 0.93
1 0.59
2 0.65
};
\addplot [draw=darkgray141160203, fill=darkgray141160203, forget plot, mark=square*, only marks]
table{%
x  y
0 0.88
1 0.68
2 0.55
};
\addplot [
  draw=yellowgreen16621684,
  fill=yellowgreen16621684,
  forget plot,
  mark=square*,
  only marks
]
table{%
x  y
0 0.74
1 0.64
2 0.41
};
\addplot [draw=burlywood229196148, fill=burlywood229196148, forget plot, mark=square*, only marks]
table{%
x  y
0 0.71
1 0.52
2 0.54
};
\addplot [semithick, mediumaquamarine102194165, dashed]
table {%
0 0.93
1 0.6
2 0.72
};
\addlegendentry{torso back}
\addplot [semithick, darkgray141160203, dashed]
table {%
0 0.86
1 0.67
2 0.67
};
\addlegendentry{torso front}
\addplot [semithick, yellowgreen16621684, dashed]
table {%
0 0.68
1 0.56
2 0.22
};
\addlegendentry{side right}
\addplot [semithick, burlywood229196148, dashed]
table {%
0 0.53
1 0.31
2 0.54
};
\addlegendentry{side left}
\addplot [semithick, mediumaquamarine102194165, dashed, forget plot]
table {%
0 0.93
1 0.59
2 0.65
};
\addplot [semithick, darkgray141160203, dashed, forget plot]
table {%
0 0.88
1 0.68
2 0.55
};
\addplot [semithick, yellowgreen16621684, dashed, forget plot]
table {%
0 0.74
1 0.64
2 0.41
};
\addplot [semithick, burlywood229196148, dashed, forget plot]
table {%
0 0.71
1 0.52
2 0.54
};
\end{axis}

\end{tikzpicture}

%% file: figures/algorithm_analysis/old_regressor_results.tex
% This file was created with tikzplotlib v0.10.1.
\begin{tikzpicture}

\definecolor{burlywood229196148}{RGB}{229,196,148}
\definecolor{darkgray141160203}{RGB}{141,160,203}
\definecolor{darkgray176}{RGB}{176,176,176}
\definecolor{lightgray204}{RGB}{204,204,204}
\definecolor{mediumaquamarine102194165}{RGB}{102,194,165}
\definecolor{yellowgreen16621684}{RGB}{166,216,84}

\begin{axis}[
legend cell align={left},
legend style={fill opacity=0.8, draw opacity=1, text opacity=1, draw=lightgray204},
tick align=outside,
tick pos=left,
x grid style={darkgray176},
xmin=-0.1, xmax=2.1,
xtick style={color=black},
xtick={0,1,2},
xticklabels={Maximum,Mean,Medium},
y grid style={darkgray176},
ylabel={Value},
ymin=-0.0995, ymax=2.5295,
ytick style={color=black}
]
\addplot [
  draw=mediumaquamarine102194165,
  fill=mediumaquamarine102194165,
  forget plot,
  mark=*,
  only marks
]
table{%
x  y
0 0.37
1 0.05
2 0.02
};
\addplot [draw=darkgray141160203, fill=darkgray141160203, forget plot, mark=*, only marks]
table{%
x  y
0 0.79
1 0.09
2 0.05
};
\addplot [draw=yellowgreen16621684, fill=yellowgreen16621684, forget plot, mark=*, only marks]
table{%
x  y
0 1.79
1 0.73
2 0.6
};
\addplot [draw=burlywood229196148, fill=burlywood229196148, forget plot, mark=*, only marks]
table{%
x  y
0 1.02
1 0.55
2 0.61
};
\addplot [
  draw=mediumaquamarine102194165,
  fill=mediumaquamarine102194165,
  forget plot,
  mark=square*,
  only marks
]
table{%
x  y
0 1.29
1 0.13
2 0.06
};
\addplot [draw=darkgray141160203, fill=darkgray141160203, forget plot, mark=square*, only marks]
table{%
x  y
0 1.02
1 0.11
2 0.07
};
\addplot [
  draw=yellowgreen16621684,
  fill=yellowgreen16621684,
  forget plot,
  mark=square*,
  only marks
]
table{%
x  y
0 2.32
1 0.05
2 0.36
};
\addplot [draw=burlywood229196148, fill=burlywood229196148, forget plot, mark=square*, only marks]
table{%
x  y
0 2.41
1 0.44
2 0.4
};
\addplot [semithick, mediumaquamarine102194165, dashed]
table {%
0 0.37
1 0.05
2 0.02
};
\addlegendentry{torso back}
\addplot [semithick, darkgray141160203, dashed]
table {%
0 0.79
1 0.09
2 0.05
};
\addlegendentry{torso front}
\addplot [semithick, yellowgreen16621684, dashed]
table {%
0 1.79
1 0.73
2 0.6
};
\addlegendentry{side right}
\addplot [semithick, burlywood229196148, dashed]
table {%
0 1.02
1 0.55
2 0.61
};
\addlegendentry{side left}
\addplot [semithick, mediumaquamarine102194165, dashed, forget plot]
table {%
0 1.29
1 0.13
2 0.06
};
\addplot [semithick, darkgray141160203, dashed, forget plot]
table {%
0 1.02
1 0.11
2 0.07
};
\addplot [semithick, yellowgreen16621684, dashed, forget plot]
table {%
0 2.32
1 0.05
2 0.36
};
\addplot [semithick, burlywood229196148, dashed, forget plot]
table {%
0 2.41
1 0.44
2 0.4
};
\end{axis}

\end{tikzpicture}

%% file: sections/fine-tuned.tex
\section{Robustifying Nominal Algorithm to Omnidirectional Faults}
\label{sec: fine_tuning_methodology}
\eva{need to shorten this section}
\gok{This section explores strategies to enhance the robustness of the nominal fall prediction algorithm against omnidirectional faults, particularly those arising from frontal-plane forces. The nominal algorithm is adaptable enough to learn new fault patterns beyond its original training. We show that this adaptability can be effectively leveraged through simple fine-tuning: updating the critical fault classifier, lead time classifier, and lead time regressor with a limited amount of additional data. This process enables the algorithm to generalize to omnidirectional faults and significantly improves performance, without requiring large-scale retraining.} \eva{Reading through the paper, I thought it was better to incorporate the adaptability portion to Section IV. Take a look and let me know what you think.}

In response to the nominal fall prediction algorithm's suboptimal performance under omnidirectional faults, particularly to faults introduced by forces applied to the frontal plane, this section explores strategies to enhance its robustness against such faults. A common challenge with learning-based methods is their reliance on large datasets. To mitigate this while leveraging the existing algorithm, we focus on fine-tuning the existing critical fault classifier, lead time classifier, and lead time regressor. Our goal is to improve performance while minimizing the need for additional training data, thus striking a balance between performance and data efficiency.

Fine-tuning is a widely used machine learning technique that adapts a pretrained model to new data by leveraging previously learned features while adjusting to the specific requirements of the new task. This approach is both efficient and effective, especially when the new data closely resembles the original data used for pretraining. Compared to training a model from scratch, fine-tuning enables rapid adaptation to new tasks with significantly less computational effort \cite{howard2018universallanguagemodelfinetuning}. 
 
For our task of standing, it is reasonable to assume that detecting faults from omnidirectional pushes shares enough similarity with the nominal algorithm's training scenario, which involves detecting faults from pushes applied to the center-back of the robot's torso. In fact, when we compared the performance of the fine-tuned model to that of a model trained from scratch, we found their performance to be comparable. 

\subsection{Fine-Tuned Critical Fault Classifier Results}
\label{sec: robust_classifier}
% \eva{Fine-tuned 5 epochs on side center pushes (middle one)}
Given the nominal critical fault classifier's poor performance on unsafe side trajectories, as discussed in Section \ref{sec: limitations_of_classifier}, it is fine-tuned only with 200 non-angled abrupt and 200 non-angled incipient faults, each applied to region B of side left and right, as shown in Fig. \ref{fig:digit_push_locations}. These regions are selected because they represent the "midpoint" of the possible pushing areas on the sides.

Despite being fine-tuned exclusively with non-angled data from region B, the fine-tuned critical fault classifier performs significantly better than the nominal critical fault classifier across all metrics, even on unseen angled-faults. Specifically, as shown in Table \ref{tab:classifier_on_normal_and_angled_pushes}, decreases significantly - from 0.08 to 0.0 for non-angled faults, and from 0.08 to 0.003 for angled faults. As expected, due to the direct relationship between false positive rate and lead time, the average lead time decreases from 2.54s to 1.18s and from 2.41s to 1.26s for non-angled and angled critical faults, respectively. On the other hand, the average response time increases by 0.16s and 0.17s, respectively, for non-angled and angled critical faults. 

When evaluated on in-distribution data, the fine-tuned model achieves a 0.0 average false positive rate, similar to the nominal model. However, the average lead time and response time decrease and increase by 0.34s and 0.18s, respectively. Despite this, the fine-tuned critical fault classifier model achieves lead times significantly above 0.2s and nearly 0 false positive rates for all critical faults, meeting the objective outlined in \cite{ref:mungai2024fall} of detecting critical faults with sufficient lead time.

%%%%%%%%%%%%%%%%%%%%%%%%%%%%%%%%%%%%%%%%%%%%%%%%%%%%%%%%%%%%%%%%%%%%%%%%%%%%%%%

\subsection{Fine-Tuned Lead Time Classifier Results}
The lead time classifier, fine-tuned using the limited set of non-angled side data as the critical fault classifier, also shows notable improvements for all faults. As shown in Table \ref{tab:lead_time_classifier_results_on_normal_and_angled_pushes}, its accuracy increases from 0.82 to 0.96 and from 0.84 to 0.95 for the non-angled and angled critical faults in the $[0,1]$ interval, respectively. The accuracy also increases from 0.57 to 0.84 and from 0.63 to 0.80 for the $(1,2]$ interval. However, the accuracy decreases from 0.61 to 0.30 and from 0.55 to 0.27, respectively, for the $(2,H]$. This decline is expected and acceptable given that the number of data points decreases exponentially as lead time increases \cite{ref:mungai2024fall}. 

When evaluated on in-distribution data, the fine-tuned model's performance slightly decreases. The accuracies drop by 0.02, 0.13, and 0.3 for the $[0,1]$, $(1,2]$, and $(2,H]$ intervals, respectively. While the drop in performance for the $(2,H]$ interval might seem significant, it is justifiable due to the previously mentioned exponential relationship between lead time and data availability. Furthermore, given the long lead times in this interval, exceeding the 0.2s threshold, it is reasonable to disregard the classifier's output for this region, as it will not have severe consequences before transitioning to intervals with more accurate predictions. 
%%%%%%%%%%%%%%%%%%%%%%%%%%%%%%%%%%%%%%%%%%%%%%%%%%%%%%%%%%%%%%%%%%%%%%%%%%%%%%%
\subsection{Fine-Tuned Lead Time Regressor Results}
Fine-tuning the lead time regressor using the same data as the critical fault and lead time classifier results in only acceptable performance for non-angled critical faults. Therefore, the lead time regressor is instead fine-tuned with 200 angled abrupt faults and 200 angled incipient faults, applied to the following regions: side left B and C, side right B and C, and torso front C. When evaluated on the in-distribution data, the maximum, mean, and median differences, as shown in Table \ref{tab:lead_time_regressor_results_on_normal_and_angled_pushes}, increase by 0.09s, 0.03s, and 0.02s, respectively, compared to the nominal model. However, when assessed on non-angled critical faults, the maximum, mean, and median differences decrease from 1.79s to 0.23s, from 0.23s to 0.03s, and from 0.07s to 0.02s, respectively, for non-angled critical faults. Similarly, for angled critical faults, the differences decrease from 2.41s to 0.32s, from 0.25s to 0.03s, and from 0.08s to 0.02s. Despite the substantial improvement in maximum difference for both angled and non-angled critical faults, the maximum difference of 0.32s achieved for angled critical faults may appear significant within the $[0,1]$ interval. However, the classifier attains this maximum difference in only 1 percent of the unsafe trajectories. This, along with the notable decrease in maximum difference, underscores the lead time regressor fine-tuned model's capability to perform well across a wide range of critical faults. 

%% file: sections/conclusion.tex
\section{Conclusion}
In this paper, we extended the nominal fall prediction algorithm from \cite{ref:mungai2024fall} to a real-time setting, implementing it in both hardware and simulation. The online implementation yields results similar to the offline version when tested on comparable data. Additionally, in simulation, it achieves an average recovery rate of 0.97. The average runtime in simulation and hardware is approximately 0.0045s and 0.0061s, respectively. The difference of 0.0061s is deemed negligible given the algorithm's sampling rate of 0.025s (40Hz). Given the consistent performance across online and offline simulations, as well as between online and offline hardware tests, we conclude that offline results are reliable indicators of online performance. Furthermore, the similar results observed between offline simulation and hardware implementations \cite{ref:mungai2024fall}, along with the comparable runtimes between online hardware and simulation implementations, suggest that the nominal fall prediction algorithm's simulation results can effectively predict its hardware outcomes.

To evaluate the nominal fall prediction algorithm’s adaptability to new scenarios, we assessed its robustness against a broader range of faults. This evaluation was conducted for each of its three components: the critical fault classifier, the lead time classifier, and the lead time regressor. All three components exhibited significantly degraded performance under these conditions, particularly for faults in the frontal plane, highlighting key limitations of the nominal algorithm.

To address these limitations while minimizing the need for additional data, we fine-tuned the three components with minimal additional data. The fine-tuned components significantly outperformed the nominal components across all evaluation metrics. Furthermore, the fine-tuned algorithm achieved an average lead time greater than 0.2s, with nearly zero average false positive rate, and a maximum average mean difference of 0.03s for the lead time. These results highlight the fine-tuned fall prediction algorithm's suitability for real-time and offline fall and lead time prediction in bipedal robots, even across a wide variety of faults.

%% file: references.bib
@article{ref:subburaman,
  title={Human inspired fall prediction method for humanoid robots},
  author={Subburaman, Rajesh and Kanoulas, Dimitrios and Muratore, Luca and Tsagarakis, Nikos G and Lee, Jinoh},
  journal={Robotics and Autonomous Systems},
  volume={121},
  pages={103257},
  year={2019},
  publisher={Elsevier}
}

@misc{ref:digit_dataset,
  author = {Mungai, M Eva and Prabhakaran, Gokul and Grizzle, Jessy W},
  title = {Digit Fall Prediction Dataset},
  year = {2023},
  publisher = {GitHub},
  journal = {GitHub repository},
  howpublished = {\url{https://github.com/UMich-BipedLab/Digit_Fall_Prediction_Dataset}},
}

@misc{ref:ar_digit,
    author = "{Agility Robotics}",
    title = "Robots",
    howpublished  = {\url{https://agilityrobotics.com/products}},
    addendum = "(accessed: 11.16.2024)"
}

@article{ref:ISERMANN19827,
title = {Process Fault Detection Based on Modeling and Estimation Methods},
journal = {IFAC Proceedings Volumes},
volume = {15},
number = {4},
pages = {7-30},
year = {1982},
note = {6th IFAC Symposium on Identification and System Parameter Estimation, Washington USA, 7-11 June},
issn = {1474-6670},
doi = {https://doi.org/10.1016/S1474-6670(17)62959-8},
url = {https://www.sciencedirect.com/science/article/pii/S1474667017629598},
author = {R. Isermann}
}

@article{ref:safaeipour2021survey,
title = {A survey and classification of incipient fault diagnosis approaches},
journal = {Journal of Process Control},
volume = {97},
pages = {1-16},
year = {2021},
issn = {0959-1524},
doi = {https://doi.org/10.1016/j.jprocont.2020.11.005},
url = {https://www.sciencedirect.com/science/article/pii/S0959152420303206},
author = {H. Safaeipour and M. Forouzanfar and A. Casavola},
}

@article{ref:wu2021falling,
  title={Falling prediction based on machine learning for biped robots},
  author={Wu, Tong and Yu, Zhangguo and Chen, Xuechao and Dong, Chencheng and Gao, Zhifa and Huang, Qiang},
  journal={Journal of Intelligent \& Robotic Systems},
  volume={103},
  pages={1--14},
  year={2021},
  publisher={Springer}
}

@phdthesis{ref:grant, title={Terrain-Aware Bipedal Locomotion}, author={Gibson, Grant}, school={University of Michigan}, year={2023} }

@article{ref:subburaman2023survey,
  title={A survey on control of humanoid fall over},
  author={Subburaman, Rajesh and Kanoulas, Dimitrios and Tsagarakis, Nikos and Lee, Jinoh},
  journal={Robotics and Autonomous Systems},
  volume={166},
  pages={104443},
  year={2023},
  publisher={Elsevier}
}

@INPROCEEDINGS{ref:mungai2023optimizing,
  author={Mungai, M. Eva and Grizzle, Jessy},
  booktitle={2023 3rd International Conference on Electrical, Computer, Communications and Mechatronics Engineering (ICECCME)}, 
  title={Optimizing Lead Time in Fall Detection for a Planar Bipedal Robot}, 
  year={2023},
  volume={},
  number={},
  pages={1-7},
  doi={10.1109/ICECCME57830.2023.10253317}}

@inproceedings{ref:mungai2024fall,
  title={Fall prediction for bipedal robots: The standing phase},
  author={Mungai, M Eva and Prabhakaran, Gokul and Grizzle, Jessy W},
  booktitle={2024 IEEE International Conference on Robotics and Automation (ICRA)},
  pages={13135--13141},
  year={2024},
  organization={IEEE}
}

@ARTICLE{ref:tran,
  author={Tran, Duy Hoa and Hamker, Fred and Nassour, John},
  journal={IEEE Transactions on Systems, Man, and Cybernetics: Systems}, 
  title={A Humanoid Robot Learns to Recover Perturbation During Swinging Motion}, 
  year={2020},
  volume={50},
  number={10},
  pages={3701-3712},
  doi={10.1109/TSMC.2018.2884619}}

@inproceedings{ref:ros,
  title={ROS: an open-source Robot Operating System},
  author={Quigley, Morgan and Conley, Ken and Gerkey, Brian and Faust, Josh and Foote, Tully and Leibs, Jeremy and Wheeler, Rob and Ng, Andrew Y and others},
  booktitle={ICRA workshop on open source software},
  volume={3},
  number={3.2},
  pages={5},
  year={2009},
  organization={Kobe, Japan}
}

@article{ref:new_igual2024time,
  title={Time Series Classification for Predicting Biped Robot Step Viability},
  author={Igual, Jorge and Parik-Americano, Pedro and Becman, Eric Cito and Forner-Cordero, Arturo},
  journal={Sensors},
  volume={24},
  number={22},
  pages={7107},
  year={2024},
  publisher={MDPI}
}

@inproceedings{ref:new_zhang2024fall,
  title={Fall Recovery Strategies in Humanoid Robots: A Brief Survey},
  author={Zhang, Boya and Jia, Ming},
  booktitle={2023 8th International Conference on Control, Robotics and Cybernetics (CRC)},
  pages={68--72},
  year={2024},
  organization={IEEE}
}

@article{ref:chevallereau2003rabbit,
  title={Rabbit: A testbed for advanced control theory},
  author={Chevallereau, Christine and Abba, Gabriel and Aoustin, Yannick and Plestan, Franck and Westervelt, Eric and De Wit, Carlos Canudas and Grizzle, Jessy},
  journal={IEEE Control Systems Magazine},
  volume={23},
  number={5},
  pages={57--79},
  year={2003}
}

@article{ref:walkman,
  title={Walk-man: A high-performance humanoid platform for realistic environments},
  author={Tsagarakis, Nikolaos G and Caldwell, Darwin G and Negrello, Francesca and Choi, Wooseok and Baccelliere, Lorenzo and Loc, Vo-Gia and Noorden, J and Muratore, Luca and Margan, Alessio and Cardellino, Alberto and others},
  journal={Journal of Field Robotics},
  volume={34},
  number={7},
  pages={1225--1259},
  year={2017},
  publisher={Wiley Online Library}
}

@misc{ref:naoh25,
    author    = {Aldebaran},
    title     = {H25 - Construction},
    howpublished = {\url{http://doc.aldebaran.com/2-1/family/nao_h25/dimensions_h25.html}},
    note      = {Accessed: 11.16.2024}
}

@misc{ref:nao,
    author = {Aldebaran},
    title = {Nao},
    howpublished = {\url{https://corporate-internal-prod.aldebaran.com/en/nao}},
    note = {Accessed: 11.16.2024}
}

@misc{ref:morgan_stanley_report,
    author = {{Morgan Stanley}},
    title = {The Humanoid 100: Mapping the Humanoid Robot Value Chain},
    howpublished = {\url{https://advisor.morganstanley.com/john.howard/documents/field/j/jo/john-howard/The_Humanoid_100_-_Mapping_the_Humanoid_Robot_Value_Chain.pdf}},
    year = {2025},
    note = {Accessed: 09.13.2025}
}

@misc{ref:wall_street_journal,
    author = {Yan Zhuang},
    title = {The Athletes at China’s Robot Games Fell Down a Lot},
    howpublished = {\url{https://www.nytimes.com/2025/08/18/world/asia/china-humanoid-robot-games.html}},
    year = {2025},
    note = {Accessed: 09.13.2025}
}

@article{DBLP:journals/corr/HendrycksG16c,
  author       = {Dan Hendrycks and
                  Kevin Gimpel},
  title        = {A Baseline for Detecting Misclassified and Out-of-Distribution Examples
                  in Neural Networks},
  journal      = {CoRR},
  volume       = {abs/1610.02136},
  year         = {2016},
  url          = {http://arxiv.org/abs/1610.02136},
  eprinttype    = {arXiv},
  eprint       = {1610.02136},
  bibsource    = {dblp computer science bibliography, https://dblp.org}
}

@misc{howard2018universallanguagemodelfinetuning,
      title={Universal Language Model Fine-tuning for Text Classification}, 
      author={Jeremy Howard and Sebastian Ruder},
      year={2018},
      eprint={1801.06146},
      archivePrefix={arXiv},
      primaryClass={cs.CL},
      url={https://arxiv.org/abs/1801.06146}, 
}

@InProceedings{ref:khaled_2025,
author="Saleh, Khaled
and O'Brien, Thomas
and Sims, Ysobel
and Mendes, Alexandre
and Chalup, Stephan",
editor="Barros, Edna
and Hanna, Josiah P.
and Okada, Hiroyuki
and Torta, Elena",
title="Efficient Sequence Model for Early Fall Detection of Humanoid Robots",
booktitle="RoboCup 2024: Robot World Cup XXVII",
year="2025",
publisher="Springer Nature Switzerland",
address="Cham",
pages="247--256",
isbn="978-3-031-85859-8"
}

@article{ref:ZHANG2025104995,
title = {Fall analysis and prediction for humanoids},
journal = {Robotics and Autonomous Systems},
volume = {190},
pages = {104995},
year = {2025},
issn = {0921-8890},
doi = {https://doi.org/10.1016/j.robot.2025.104995},
url = {https://www.sciencedirect.com/science/article/pii/S0921889025000818},
author = {Chiyu Zhang and Jie Gao and Ziyu Chen and Shanlin Zhong and Hong Qiao}
}

@inproceedings{ref:gong2019feedback,
  title={Feedback control of a cassie bipedal robot: Walking, standing, and riding a segway},
  author={Gong, Yukai and Hartley, Ross and Da, Xingye and Hereid, Ayonga and Harib, Omar and Huang, Jiunn-Kai and Grizzle, Jessy},
  booktitle={2019 American control conference (ACC)},
  pages={4559--4566},
  year={2019},
  organization={IEEE}
}
